\documentclass{article}
\usepackage{ijcai24}

\usepackage{times}
\usepackage{soul}
\usepackage{url}
\usepackage[hidelinks]{hyperref}
\usepackage[utf8]{inputenc}
\usepackage[small]{caption}
\usepackage{graphicx}
\usepackage{amsmath}
\usepackage{amsthm}
\usepackage{booktabs}
\usepackage[switch]{lineno}

\usepackage{helvet}  
\usepackage{courier}  
\usepackage{amsmath,amssymb,amsfonts,amssymb,multirow}
\usepackage{graphicx}
\usepackage{xcolor}
\usepackage{subfigure}
\usepackage{threeparttable}
\usepackage{url,setspace}
\usepackage{colortbl}
\usepackage[boxed,ruled,vlined,linesnumbered]{algorithm2e}
\usepackage{algpseudocode}
\usepackage{multirow}
\usepackage{multicol}
\usepackage{makecell}
\usepackage{textcomp}
\usepackage{enumitem}
\usepackage{booktabs}
\usepackage{tablefootnote}
\usepackage{lineno}

\usepackage{amsmath,amsfonts,bm}

\def\eqref#1{equation~\ref{#1}}

\def\1{\bm{1}}

\def\vx{{\bm{x}}}
\def\vy{{\bm{y}}}
\def\vz{{\bm{z}}}

\def\mM{{\bm{M}}}

\def\mX{{\bm{X}}}

\DeclareMathAlphabet{\mathsfit}{\encodingdefault}{\sfdefault}{m}{sl}
\SetMathAlphabet{\mathsfit}{bold}{\encodingdefault}{\sfdefault}{bx}{n}

\DeclareMathOperator*{\argmin}{arg\,min}

\usepackage{cleveref}

\SetKwInOut{Parameter}{Parameter}

\let\oldnl\nl
\newcommand{\nonl}{\renewcommand{\nl}{\let\nl\oldnl}}
\newcommand{\ie}{\textit{i}.\textit{e}.}
\newcommand{\eg}{\textit{e}.\textit{g}.}

\title{Trade When Opportunity Comes: Price Movement Forecasting via Locality-Aware Attention and Iterative Refinement Labeling}

\author{
Liang Zeng$\textsuperscript{\textmd{1}}$
\and Lei Wang$\textsuperscript{\textmd{1}}$
\and Hui Niu$\textsuperscript{\textmd{1}}$
\and Ruchen Zhang$\textsuperscript{\textmd{2}}$ 
\and Ling Wang$\textsuperscript{\textmd{2}}$
\textnormal{and} Jian Li$\textsuperscript{\textmd{1}}$
\\
\affiliations
$\textsuperscript{1}$Institute for Interdisciplinary Information Sciences (IIIS), Tsinghua University, China     
\\
$\textsuperscript{2}$Huatai Securities Co., Ltd, China    
\\
\{zengl18, wanglei20, niuh17\}@mails.tsinghua.edu.cn; \{zhangruchen, wangling\}@htsc.com \\
lijian83@mail.tsinghua.edu.cn
}

\begin{document}

\maketitle
\begin{abstract}
Price movement forecasting, aimed at predicting financial asset trends based on current market information, has achieved promising advancements through machine learning~(ML) methods.
Most existing ML methods, however, struggle with the extremely low signal-to-noise ratio and stochastic nature of financial data, often mistaking noises for real trading signals without careful selection of potentially profitable samples.
To address this issue, we propose LARA, 
a novel price movement forecasting framework with two main components: \underline{L}ocality-Aware \underline{A}ttention~(LA-Attention) and Iterative \underline{R}efinement L\underline{a}beling~(RA-Labeling). 
(1) LA-Attention, enhanced by metric learning techniques, automatically extracts the potentially profitable samples through masked attention scheme and task-specific distance metrics.
(2) RA-Labeling further iteratively refines the noisy labels of potentially profitable samples, and combines the learned predictors robust to the unseen and noisy samples. 
In a set of experiments on three real-world financial markets: stocks, cryptocurrencies, and ETFs, LARA significantly outperforms several machine learning based methods on the Qlib quantitative investment platform. Extensive ablation studies confirm LARA's superior ability in capturing more reliable trading opportunities.

\end{abstract}
\section{Introduction}
\label{sec:introduction}
Price movement forecasting, a crucial yet  challenging task in quantitative finance, is notoriously difficult largely due to the financial market's inherently stochastic, dynamic, and volatile nature~\cite{de2018advances}.
The classical Efficient Market Hypothesis (EMH)~\cite{fama1960efficient,lo2019adaptive} states that in an informationally efficient market, price changes must be unforecastable if all relevant information is reflected immediately.
However, perfect efficiency is impossible to achieve in practice
\cite{campbell1997econometrics} due to informational asymmetry and ``noisy'' behaviors among the traders \cite{bloomfield2009noise}.
Indeed, hundreds of abnormal returns have been discovered in the literature~(see the classical review by~\cite{grossman1980impossibility} and the recent list \cite{harvey2016and}).
As the more recent endeavor to ``beat the market'' and achieve excess returns, machine learning based solutions, taking advantage of the nonlinear expressive power, have become increasingly popular and achieved promising results~\cite{de2018advances,gu2018empirical,zhang2020doubleensemble,xu2020adaptive,wang2019clvsa}.

\begin{figure}[tp]
	\centering
	\includegraphics[width=.48\textwidth]{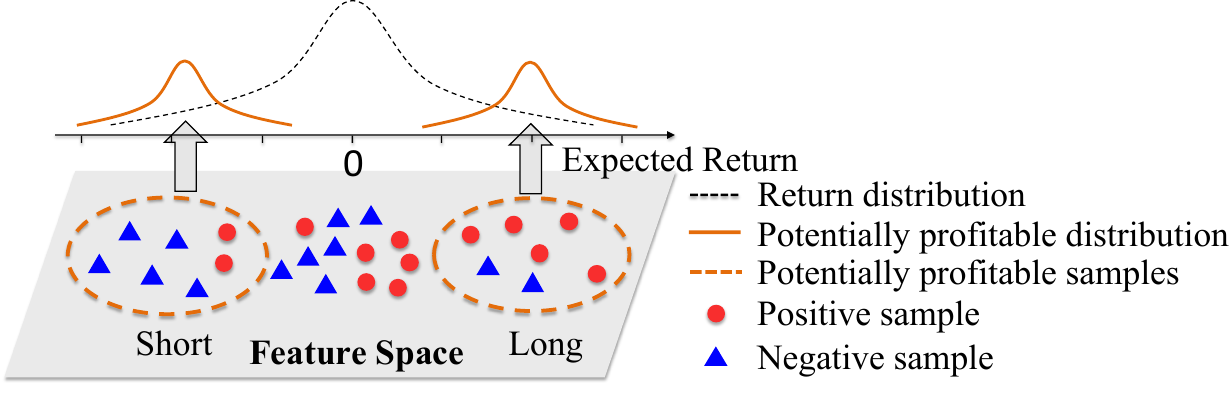}
	\caption{Illustration of the potential profitable samples. The half top figure represents the probability density function~(PDF) of expected return over corresponding samples~(best viewed in color).}
	\label{fig:intuition}
\end{figure}

The straightforward approach to utilizing machine learning techniques for financial forecasting can be distilled into two primary steps \cite{gu2018empirical}.
First, mine effective factors (either manually or automatically~\cite{li2019individualized}) correlated/predictive with asset returns.
Second, feed these factors as features into some off-the-shelf machine learning algorithms to generate trading signals. Recently, much effort has been devoted to designing new machine learning algorithms in the second step to make more accurate predictions~\cite{zhang2020doubleensemble,xu2020adaptive,zhang2017stock,wang2019alphastock,yang2020qlib,de2018advances}. 
Most works mentioned above trains the model over the entire set of training data (typically spans a consecutive period of time)\footnote{Note that in financial time-series forecasting problems, one should not split the training and testing set randomly, since the data points are not i.i.d., but strongly and temporally correlated.}. 
However, it is well known that financial time-series data has an extremely low signal-to-noise ratio~\cite{de2018advances}, to the degree that modern machine algorithms are prone to pick up patterns of noise instead of real profitable signals (\emph{a.k.a.,} overfit the training set).
As depicted in Fig.~\ref{fig:intuition}, samples are mixed together in the feature space, making it challenging to clearly distinguish between them.
Without careful selection of potentially profitable samples from the entire set of training data, tuning the models to generalize well can be very challenging~\cite{zhang2017stock,kim2021fine}. Even for the most experienced investment experts, asset prices often exhibit behaviors akin to unpredictable random walks, with profitable opportunities occurring only sporadically.
In light of this fact, we argue that it can be more effective and robust to focus on specific samples that are potentially more predictable and profitable~(Fig.~\ref{fig:intuition}), as the proverb says, \emph{trade when opportunity comes}. Consequently, our goal is to develop a powerful and generalizable model for price movement forecasting over highly noisy data. However, this realistic financial scenario brings two unique technical challenges, which we will tackle in this paper.

\textbf{Challenge 1: How to effectively extract potentially profitable samples?}

\textbf{Solution 1:} \emph{Locality-aware attention}~(LA-Attention).
We design two main modules for LA-Attention: a \emph{metric learning module} for constructing a better metric, and a \emph{localization module} for explicitly extracting potentially profitable samples. 
Specifically, many algorithms rely critically on being given a good metric over their inputs due to the complex internal relations among different features of financial data~\cite{xing2002distance}.
It motivates us to introduce the \emph{metric learning module} to assist the following \emph{localization module} by learning a more compatible distance metric~\cite{de2020metric}. 
The \emph{localization module} intends to model an auxiliary relation between input samples and their labels by locally integrating label information to extract the desired potentially profitable samples.

\textbf{Challenge 2: How to adaptively refine the labels of potentially profitable samples?}

\textbf{Solution 2:} \emph{Iterative refinement labeling}~(RA-Labeling). We empirically identify that in the actual financial market, two samples may have similar features but yield completely opposite labels because of the stochastic and chaotic nature of financial data~\cite{de2018advances,lo2011non}.
Such \emph{noisy samples}, \textit{i.e.,} similar samples with the opposite labels, hamper the model generalization in practical financial applications~\cite{song2022learning}.
In response, we propose the novel RA-Labeling method to adaptively distinguish noisy labels of potentially profitable samples according to their training losses, iteratively refine their labels with multiple predictors, and combine the learned predictors to enhance the overall performance.

\begin{figure*}[htp]
    \centering
    \includegraphics[width=17.5cm]{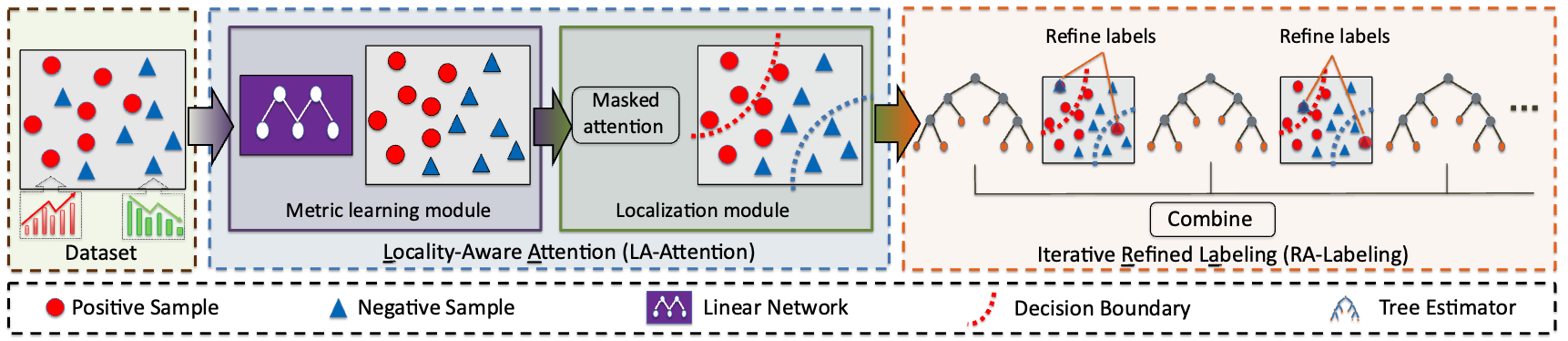}
    \caption{The workflow of our proposed LARA framework. LARA first extracts potentially profitable samples from the noisy market and then refines their labels. LARA consists of two sequential components: LA-Attention and RA-Labeling.}
    \label{fig:framework}
\end{figure*}

In short, our main \textbf{contributions} are as follows:
\begin{itemize}[leftmargin=*]

    \item We propose the two-step LA-Attention method to first locally extract potentially profitable samples by attending to label information and then construct a more accurate classifier based on these selected samples. Moreover, we utilize a metric learning module to learn a well-suited distance metric to assist LA-Attention.
    \item We introduce the RA-Labeling method to adaptively refine the noisy labels of potentially profitable samples. We obtain a noise-robust model with our proposed combination schemes to integrate the multiple learned predictors and boost generalization ability.
    \item We conduct comprehensive experiments on three real-world financial markets. LARA significantly outperforms a set of machine learning competitors on the Qlib quantitative investment platform~\cite{yang2020qlib}. Extensive ablation studies and experiments demonstrate the effectiveness of each component in LARA and suggest that LARA indeed captures more reliable trading opportunities.
\end{itemize}


\section{Related Work}
\label{sec:related}
\paragraph{Price Movement Forecasting.} 
There are two main research threads in price movement forecasting based on machine learning methods. 
(1) \cite{li2019individualized,achelis2001technical,sawhney2021exploring,xu2020adaptive} focus on improving prediction by introducing extra financial data sources. \cite{gu2018empirical} used sentiment analysis to extract supplementary financial information from various textual data sources.
Besides, there also exists much literature~(\eg,~\cite{li2019individualized,achelis2001technical}) studying new methods to construct more informative features for financial data.
(2) \cite{zhang2020doubleensemble,wu2021temp,jiang2022adamct,zhang2017stock} aim to develop specialized prediction models to enhance prediction accuracy. 
\cite{zhang2020doubleensemble} proposed a new ensemble method based on the sample reweighting and feature selection method for stock price prediction.
Despite increasing efforts in price movement forecasting, few of them explicitly incorporate auxiliary label information in an extremely noisy market.
We propose the two-step LA-Attention to effectively extract potentially profitable samples.

\paragraph{Handling Label Noise.} 
Learning with noisy labels is a common challenge in many real-world applications~\cite{bloomfield2009noise,zhang2020doubleensemble,song2022learning}.
\cite{chen2021beyond} proposed a new labeling method called SEAL~(Self-Evolution Average Label) to tackle instance-dependent label noise. 
Apart from the labeling method, \cite{kim2021fine} proposed filtering noisy instances via their eigenvectors~(FINE), while \cite{xia2021sample} dynamically selected training samples by analyzing the prediction values between two models.
However, in the context of quantitative investment, these methods may not suffice due to extreme noise levels, as illustrated in Fig.~\ref{fig:intuition}. 
Simple label modifications or noisy sample removal can inadvertently increase uncertainty or eliminate the potentially profitable opportunities.
In contrast, our RA-Labeling method is designed to adaptively refine the noisy labels of potentially profitable samples to build a noise-robsut model.
\section{Preliminary}
\label{sec:preliminary}

\paragraph{Problem Setup.} Consider a time series of the asset trading price over $T$ time-steps $\{\text{Price}_t|t=1,2,\dots,T\}$. 
Typically, the price movement trend is defined as the future change~(return) of the asset price~\cite{de2018advances}. In this paper, we adopt a fixed-time horizon method to define the asset price trend. Concretely, the label $y_t$ indicates whether the return of the asset price over a time horizon $\Delta$ exceeds a certain pre-defined threshold $\lambda$. For long positions:
\begin{equation}
    y_t = 
    \left\{
    \begin{array}{cl}
        1 ,& \text{Price}_{t+\Delta} / \text{Price}_{t} - 1 > \lambda \\
        0 ,& \text{otherwise}
    \end{array}
    \right.
    .
\end{equation}
Similarly, for short positions, we set the label 1 if $\text{Price}_{t+\Delta} / \text{Price}_{t} - 1 < -\lambda$ and 0 otherwise.
At the training step $t$, we calculate the feature vector $\vx_t \in \mathcal{X} \subseteq \mathbb{R}^d$ based on price and volume information up to time $t$, and its label $y_t \in \mathcal{Y} = \{0, 1\}$.
We denote the training set as $\mX = [\vx_1, \cdots, \vx_N]^T \in \mathbb{R}^{N \times d}$ associated with its label $\vy=[y_1,\cdots,y_N]^T \in \mathbb{R}^{N}$, where $N$ is the number of time steps and $d$ is the feature dimension. Similarly, $\mX_{\text{test}} = [\vx_1, \cdots, \vx_{N_{\text{test}}}]^T \in \mathbb{R}^{N_{\text{test}} \times d}$ denotes the testing set.

\paragraph{Goal.} \emph{Price movement forecasting} aims to build a (binary) parameterized classifier $f_{\theta}:\mathcal{X} \rightarrow \mathcal{Y} $, where $f_\theta(\vx_t) = Pr(y_t=1)$ represents the probability of whether the price trend $y_t$ exceeds the pre-defined threshold $\lambda$. In order to prevent the data leakage issue of financial data, $f_\theta(\vx_t)$ makes a prediction on the asset price trend $y_t$ of $\Delta$-step ahead~($t+\Delta$) only based on price information up to $t$.

\paragraph{Samples With High \bm{$p_x$}.} 
Because we consider the binary classification, we assume the predicted label $y_{t} \sim \text{Bernoulli}(p_{\vx_t})$ conditional on known information $\vx_t$ at the $t$ time step. It means that we observe the positive class with the probability $p_{\vx_t}$ based on the current market condition.
A larger $p_{\vx_t}$ indicates a stronger tendency and lower uncertainty of the asset price, and thus the prediction at this time step can be more reliable. In order to make a more accurate prediction, we gather the \emph{samples with high $p_{\vx_t}$}, where $p_{\vx_t}$ is greater than a given threshold~$\left(thres\right)$, \emph{i.e.,} $p_{\vx_t} > thres$. 
However, it is impossible to obtain the exact underlying probability $p_{\vx_t}$ from the input data since we can only observe one sample of the distribution $y_{t} \sim \text{Bernoulli}(p_{\vx_t})$ at one certain time-step. Thus, we introduce \emph{locality-aware attention} (Sec.~\ref{sec:attention}) to give an approximate estimation $\hat{p}_{\vx_t}$ of $p_{\vx_t}$.
\section{LARA: The Proposed Framework}
\label{sec:method}
In this section, we present our LARA framework, which consists of two main components---\emph{locality-aware attention}~(LA-Attention, Sec.~\ref{sec:attention}) and \emph{iterative refinement labeling}~(RA-Labeling, Sec.~\ref{sec:bi-level}).
The key idea of LARA is to extract potentially profitable samples from the noisy market. LA-Attention extracts the \emph{samples with high $p_\vx$} and RA-Labeling refines the labels of \emph{noisy samples} to boost the performance of predictors. 
The workflow of LARA is shown in Fig.~\ref{fig:framework}.

\subsection{\underline{L}ocality-Aware \underline{A}ttention~(LA-Attention)}
\label{sec:attention}
LA-Attention is a two-step method for its training and testing phases.
In the training phase, we first extract the \emph{samples with high $p_{\vx_t}$} satisfying $\hat{p}_{\vx_t} > thres$~($thres$ denotes the desired probability that one sample yields the positive class), which implicitly denote the potentially profitable samples.
Then we train a more robust model on these selected samples according to the corresponding supervised losses. 
In the testing phase, we first gather the \emph{samples with high $p_{\vx_t}$} by the $\hat{p}_{\vx_t} > thres$ criterion, and then we only make predictions on these selected samples.
Thanks to this two-step modular design, LA-Attention can largely improve the precision of the prediction model on empirical evaluations. 
The complete algorithm is summarized in Algorithm \ref{alg:semi-lazy-training}. 
In what follows, we first introduce the localization module to estimate $\hat{p}_{\vx_t}$ and then introduce the metric learning module to assist the \emph{localization module} by learning a well-suited distance metric.

\paragraph{Localization Module.}
\label{sec:localization module}
It is hard to recover the label distribution $y_{t} \sim \text{Bernoulli}(p_{\vx_t})$ from the noisy asset price dataset, as we can only observe one sample of the distribution at the $t$-th time-step. Suppose the parameter $p_{\vx_t}$ of the probability distribution is continuous with respect to known information $\vx_t$~(\emph{e.g.,} the asset price). 
We design a \emph{localization module} attending to other observed samples in the dataset to obtain an approximate estimation $\hat{p}_{\vx_t}$ of $p_{\vx}$. In light of the concepts of attention mechanism~\cite{vaswani2017attention}, we denote $\left(\vx_i, y_i\right)$ as the key-value pair, where $y_i$ is the label of $\vx_i$. Let $\vx_t$ be the query sample. The \emph{localization module} can be formulated as follows:
\begin{equation}
    \label{eq:label-attention}
    \hat{p}_{\vx_t} = \sum_{1 \leq i \leq N} y_i \cdot \frac{k(\vx_i, \vx_t)}{\sum_{1 \leq j \leq N} k(\vx_j, \vx_t)} \quad,
\end{equation}
where $N$ is the size of training data, and $k(\vx_i, \vx_t)$ is the attention weight between $\vx_i$ and $\vx_t$. 
However, in the most general attention mechanisms, the model allows each sample to attend to every other sample.
Intuitively, closer neighbors of a query sample have greater \emph{similarity} than neighbors which are further away.
We inject the locality structure into the module by performing the \textit{masked attention} scheme and merely paying attention to the neighbors $\mathcal{N}(\vx_t)$ of $\vx_t$. Then we reformulate the \emph{localization module} as follows:
\begin{equation}
    \label{eq:label-attention-reformulate}
    \hat{p}_{\vx_t} = \sum_{\vz_i \in \mathcal{N}(\vx_t)} y_{\vz_i} \cdot \frac{k(\vz_i, \vx_t)}{\sum_{\vz_j \in \mathcal{N}(\vx_t)} k(\vz_j, \vx_t)} \quad.
\end{equation}
We propose two schemes~(detailed in Appendix) to define the neighbors $\mathcal{N}(\vx_t)$ in the \emph{localization module}: 1) the k-nearest neighbors of the query point (K-Neighbor); 2) the k-nearest neighbors within a moderate radius (R-Neighbor).
Moreover, there exist several reasonable similarity metrics to implement the attention weight in Eq.(\ref{eq:label-attention-reformulate}). We recommend two practical approaches: 1) the identical weight; 2) the reciprocal of Mahalanobis distance $d_{\mM}(\vz, \vx_t)$~(introduced below), detailed as follows:
\begin{equation}
    \label{eq:attention-weight}
    k(\vz,\vx_t) := \left\{
    \begin{array}{ll}
         k_I(\vz, \vx_t)= 1  \\
         k_R(\vz, \vx_t)= 1/d_{\mM}(\vz, \vx_t)
    \end{array}
    \right. \;.
\end{equation}

\paragraph{Metric Learning Module.}
\label{sec:metric}
As introduced above, we need a reasonable measurement of the distance between data representations on the embedding space to select \emph{samples with high $p_{x_t}$}.
However, in the context of quantitative investment, different investors have different understandings of the financial market and prefer completely different types of factors, \emph{e.g.,} value investors prefer fundamental factors, while technical analysts may choose technical factors.
Due to the multi-modal nature of financial data, we adopt the metric learning technique to mitigate the pressure of manually searching for the suitable distance measurement and enhance the flexibility of LARA.
The goal of metric learning is to learn a Mahalanobis distance metric $\mM$ to pull the similar samples close and push the dissimilar samples away. A straightforward solution is to minimize the dissimilarity of samples within the same class. We first define the indicator function as $K_{i,j}=1$ if $y_i = y_j$ and $0$ otherwise.
Given two embedding vectors $\vx_i, \vx_j \in \mathbb{R}^d$, we optimize the~(squared) Mahalanobis matrix $\mM \in \mathbb{R}^{d\times d}$~\cite{kaya2019deep} as follows:
\begin{equation}
\begin{aligned}
    &\min_{\mM} \quad \frac{1}{2} \sum \limits_{\vx_i,\vx_j \in \mathcal{X}} d_{\mM}(\vx_i,\vx_j) K_{i,j} \\
    =& \min_{\mM} \quad \frac{1}{2} \sum \limits_{\vx_i,\vx_j \in \mathcal{X}} (\vx_i-\vx_j)^T\mM(\vx_i-\vx_j) K_{i,j} \\
    =&\min_{\mM} \sum \limits_{\vx_i,\vx_j \in \mathcal{X}} (\vx_i^T\mM\vx_i - \vx_i^T\mM\vx_j) K_{i,j} \quad,
\end{aligned}
\end{equation}
where $d$ is the feature dimension, and $d_{\mM}(\vx_i,\vx_j)$ denotes the Mahalanobis distance.
Metric learning module can better leverage the similarity and dissimilarity among samples by learning a new embedding space.
Thus, LARA adopts the \emph{metric learning module} to automatically learn a well-suited distance metric in the embedding space, assisting the \emph{localization module} in extracting potentially profitable samples.

\begin{figure}[tbp]
	\centering
	\includegraphics[width=.48\textwidth]{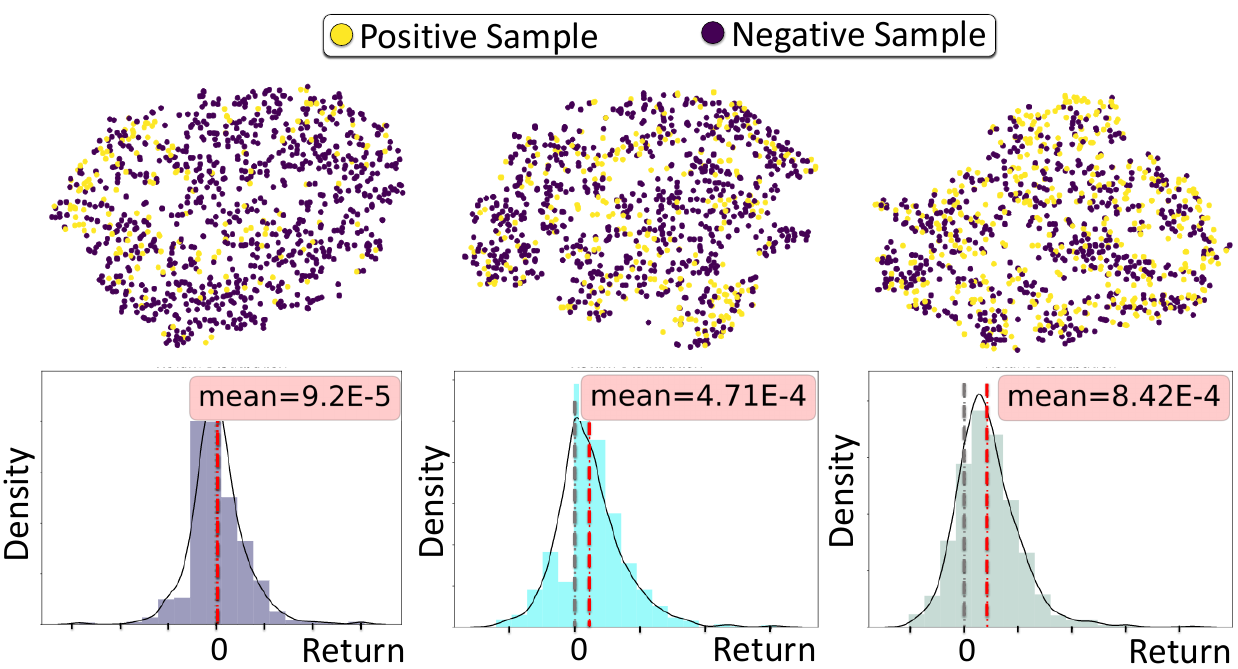}
	\caption{Visualization of randomly sampled 1000 points with t-SNE and their corresponding return distributions. \textbf{Left:} the input dataset. \textbf{Middle:} \emph{samples with high $p_{\vx_t}$} selected by \emph{localization module}. \textbf{Right:} \emph{samples with high $p_{\vx_t}$} selected by LA-Attention~(\emph{localization module} + \emph{metric learning module}).}
	\label{fig:tsne}
\end{figure}

\paragraph{LA-Attention Has a Proper Estimation of $p_{\vx_t}$.}
\label{sec:understanding}
We use a running example to illustrate why LA-Attention helps extract potentially profitable samples and further understand the effectiveness of the~\emph{metric learning module} incorporated into the LA-Attention method, as shown in Fig.~\ref{fig:tsne}. 
We search for potentially profitable~\emph{samples with high $p_{\vx_t}$} via the $\hat{p}_{\vx_t}>thres$ criterion~($thres$ equals 0.5 in this demo).
We aim to verify whether $\hat{p}_{\vx_t}$ calculated by LA-Attention is a proper estimation of the predefined $p_{\vx_t}$.
In Fig.~\ref{fig:tsne}~(Left), We plot embeddings of randomly sampled 1000 points in the ETF dataset using t-SNE \cite{van2008visualizing}. 
We find that the positive and negative points are mixed together and there are only 174 positive samples. 
It exhibits the return distribution with almost the zero mean, and $\hat{p}_{\vx_t} \approx 0.174$ is far away from the desired $p_{\vx_t}$.
After introducing the \emph{localization module} (Fig.~\ref{fig:tsne}~(Middle)), we also randomly sample 1000 points with $\hat{p}_{\vx_t}>thres$ criteria, and find that there are 279 positive samples. The mean return of potentially profitable samples is obviously larger than zero, which indicates the effectiveness of the \emph{localization module}. 
However, the negative ones still account for a large part of the samples and the estimated $\hat{p}_{\vx_t}$~(0.279) is much lower than desired $thresh=0.5$. 
In Fig.~\ref{fig:tsne}~(Right), further incorporated into the \emph{metric learning module}, the number of the positive samples (424) accounts for almost half of all samples, which is in line with expectations that $\hat{p}_{\vx_t}$ equals 0.5 and is a proper estimation of $p_{\vx_t}$.
We also find that the learned embedding has a more compact structure and clearer boundaries. 
It achieves the $8.42E\text{-}4$ mean return, which is much larger than zero with a greater margin compared with the other two methods~(note that $8.42E\text{-}4$ is a reasonable and effective return in high-frequency trading and it is close to our target return of $\lambda = 1E\text{-}3$).
Hence, this learned distance metric can assist the LA-Attention method in distinguishing potentially profitable \emph{samples with high $p_{\vx_t}$}.

\begin{algorithm}[htbp]
    \caption{LARA framework}
    \label{alg:semi-lazy-training}
    \SetCommentSty{textit}
    \SetKwComment{Comment}{}{}
    \KwIn{Training data $\left(\mX, \vy\right) \in \mathbb{R}^{N \times d} \times \mathbb{R}^{N}$, testing data $\mX_{\text{test}} \in \mathbb{R}^{N_{\text{test}} \times d}$, the model $f$. }
    \Parameter{Threshold $thres$, combination scheme $C$, hyper-parameters $K$.}
  	\KwOut{$Y_{out} \in \mathbb{R}^N$.}
        \nonl $\vartriangleright$ \textbf{Training phase: } \\
  	Init \textit{emtpy set} $\mX', \vy'$. \\
  	\For{$(\vx,y)$ \textbf{in} $(\mX, \vy)$}{
  	\nonl    Estimate $\hat{p}_\vx$ via Eq.~\ref{eq:label-attention-reformulate} with the query sample $\vx$ and the key-value pair $(\vx_i, y_i) \in (\mX,\vy)$.\\
  	    \textbf{if} $\hat{p}_x$\textgreater $thres$ \textbf{then} $\mX' \cup \{\vx\} ; \; \vy' \cup \{y\}$. \\
  	}
  	\nonl $F$ = RA-Labeling $\left( {\mX'}, {\vy'}, f, K, C\right)$ \Comment*[r]{\textcolor{blue}{/*\textit{Algorithm~\ref{alg:Bi-level-label}}*/}}
  	\nonl $\vartriangleright$ \textbf{Testing phase: } \\
  	Init \textit{emtpy set} $\mX_{\text{test}}'$. \\
  	\For{$\vx_{\text{test}}$ \textbf{in} $\mX_{\text{test}}$}{
  	\nonl    Estimate $\hat{p}_{\vx_{\text{test}}}$ via Eq.~\ref{eq:label-attention-reformulate} with the query sample $\vx_{\text{test}}$ and the key-value pair $(\vx_i, y_i) \in (\mX,\vy)$.\\
  	    \textbf{if} $\hat{p}_\vx$\textgreater $thres$ \textbf{then} $\mX_{\text{test}}' \cup \{\vx_{\text{test}}\}$. \\
  	}
  	\nonl \textbf{Return:} $Y_{out} = F(\mX_{\text{test}}')$. \\
\end{algorithm}

\subsection{Iterative \underline{R}efinement L\underline{a}beling~(RA-Labeling)}
\label{sec:bi-level}
Price movement is notoriously difficult to forecast because financial markets are complex dynamic systems~\cite{de2018advances,lo2011non}.
Indeed, we observe that the complicated financial markets sometimes mislead ML systems to produce opposite prediction results with extremely high confidence. 
For example, some training samples get a high~(low) probability of the positive prediction $Pr(y=1)$, but their labels are negative~(positive).
Such \emph{noisy samples} hamper the model generalization in price movement forecasting~\cite{livisualizing}.
Thus, LARA should be robust to noisy labels of potentially profitable samples, and we adopt the label refurbishment technique introduced by~\cite{song2022learning} to learn a noise-robust prediction model.
Formally, given the training data $\mX \in \mathbb{R}^{N \times d}$, our method runs in an iterative manner to obtain adaptively refined labels $\{\vy^{k}\}_{0 \leq k \leq K}$, where $N$ is the number of training data and $K$ is the number of rounds in iterative refinement labeling. 

We denote the trained predictor at the $k$-th round and the current prediction of the sample $j$ at the $k$-th round as $\{f_{\theta^{k}}\}_{0 \leq k \leq K}$ and $f_{\theta^{k}}(\vx_j)$, respectively.
The refurbished label $y_{j}^{k+1}$ can be obtained by a convex combination of the current label $y_{j}^{k}$ and the associated prediction $f_{\theta^k}(\vx_j)$:
\begin{equation}
    y_{j}^{k+1} = \alpha_{j}^k y_{j}^{k} + (1-\alpha_{j}^k) f_{\theta^{k}}(\vx_j),
    \label{equ:ema}
\end{equation}
where $\alpha_{j}^k \in [0,1]$ is the label confidence and $y_{j}^{0}$ is the initial noisy label of training data. As shown in Eq.~(\ref{equ:ema}), we apply the exponential moving average to mitigate the damage of under-predicted labels of \emph{noisy samples} and thus the confidence score $\alpha_{j}^k$ is a crucial parameter in this self-adaptive training scheme. We propose to update $\alpha_{j}^k$ according to the training loss at the current step to alleviate the instability issue of using instantaneous prediction under \emph{noisy samples}. $\alpha_{j}^k$ of the sample $\vx_j$ at the $k$-th step is calculated by:
\begin{equation}
    \alpha_{j}^k = 1 \left\{ \mathcal{L} \left(f_{\theta^{k}}(\vx_j), y_{j}^{k} \right) \leq \epsilon^k \right\},
\end{equation}
where $1\left\{\cdot\right\}$ denotes the indicator function, $\epsilon^k$ is a hyper-parameter to control label confidence, and $\mathcal{L}(\cdot, \cdot)$ denotes the supervised loss function, \emph{e.g.}, the cross-entropy loss. 
In order to develop a more coherent predictor that improves its ability to evaluate the consistency of \emph{noisy samples}, we adaptively set the value of $\epsilon^k$ to roughly adjust the ratio of labels in training data, denoted by $r$.
We also perform the hyperparameter study in Sec.~\ref{sec:q3} and empirically identify that RA-Labeling achieves a relatively stable performance when $r<10\%$.

Different from the conventional noisy labeling methods~\cite{song2022learning}, we further propose a combination method to integrate the learned predictor. Specifically, after running $K$ iterations of iterative refinement labeling, we obtain $K+1$ iterated predictors $\{f_{\theta_0},\cdot \cdot \cdot,f_{\theta_K}\}$. We integrate all predictors using the following two different combination schemes: (1) $C=C_{\text{Last}}$: making a prediction according to the last iterated predictor. (2) $C=C_{\text{Vote}}$: making a prediction that averages all iterated predictors. Then the output prediction model $F$ is defined as follows:
\begin{equation}
\begin{aligned}
& \quad F(\vx) := \\ 
&  
    \left\{
    \begin{aligned}
        C_{\text{Last}}\left( \{f_{\theta^k} \}_{0 \leq k \leq K} \right) (\vx) &= f_{\theta^{K}}(\vx), \qquad \quad&C=C_{\text{Last}}   \\
        C_{\text{Vote}}\left( \{f_{\theta^k} \}_{0 \leq k \leq K} \right) (\vx) &= \mathbb{E}_k \left[f_{\theta^k}(\vx)\right], &C=C_{\text{Vote}}
    \end{aligned}
    \right. .
\end{aligned}
\end{equation}
Note that RA-Labeling can be seamlessly incorporated into existing machine learning methods to mitigate the negative consequences of learning from noisy labels. The pseudocode of RA-Labeling is shown in Algorithm~\ref{alg:Bi-level-label}.

\begin{algorithm}[t]
    \caption{RA-Labeling}
    \label{alg:Bi-level-label}
    \KwIn{Training data $\left(\mX, \vy^{0}\right) \in \mathbb{R}^{N \times d} \times \mathbb{R}^{N}$, the  model $f: \mathbb{R}^{d} \rightarrow \mathbb{R}$. }
  	\Parameter{Iteration number $K$, combination scheme $C$.}
  	\KwOut{$F: \mathbb{R}^d \rightarrow \mathbb{R}$}
  	\nonl Init: $\theta^0 = \argmin_{\theta^0} \sum_{j=1}^N \mathcal{L} \left(f_{\theta^0}(\vx_j), y^{0}_j\right)$ \\
    \For{$k \in \left\{0,\dots,K-1\right\}$}{
        \nonl \For{$j \in \left\{1,\dots,N\right\}$}{
            \footnotesize
        \nonl $y_{j}^{k+1} = \alpha_{j}^k y_{j}^{k} + (1-\alpha_{j}^k) f_{\theta^{k}}(\vx_j)$
        }
        \nonl $\theta^{k+1} = \argmin_{\theta^{k+1}} \sum_{j=1}^N \mathcal{L} \left(f_{\theta^{k+1}}(\vx_j), y^{k+1}_j \right)$ \\
    }
    \nonl \textbf{Return:} $F = C\left( \{f_{\theta^k} | 0 \leq k \leq K\} \right)$. \\
\end{algorithm}
\section{Experiments}
\label{sec:experiments}

\newcommand{\rednum}[1]{\textcolor{red}{\textbf{#1}}} 
\newcommand{\bluenum}[1]{\textcolor{blue}{\underline{#1}}}
\newcommand{\orangenum}[1]{\textcolor{orange}{\textbf{#1}}}

\begin{table*}[htbp]
    \centering
    
    \scalebox{1.0}{
    \begin{tabular}{cl|ccc | ccc | cc}
    \toprule
        \multicolumn{2}{c|}{\multirow{2}*{Methods}} & \multicolumn{3}{c}{China's A-share stocks} & \multicolumn{3}{c}{Cryptocurrency} & \multicolumn{2}{c}{Ranking Count}\\ 
        \cmidrule(lr){3-5} \cmidrule(lr){6-8} \cmidrule(lr){9-10} 
        & & PR(\%) & WLR & AR & PR(\%) & WLR & AR & $1^{st}$ & $2^{nd}$\\
    \midrule
        \multirow{7}*{\begin{tabular}{c}Quantitative\\Investment\\Methods\end{tabular}}
        & \multicolumn{1}{l|}{ALSTM} & 51.9 & 1.133 & 3.24E-3 & 47.9 & 0.933 & 5.9E-5 & 0 & 0\\
        & \multicolumn{1}{l|}{TabNet} & 51.8 & 1.299 & 5.14E-3 & 51.0 & 0.890 & 1.4E-5 & 0 & 0 \\
        & \multicolumn{1}{l|}{Transformer} & 53.2 & 1.230 & 5.73E-3 & 38.7 & 0.888 & -5.0E-5\hspace{0.3em} & 0 & 0 \\
        & \multicolumn{1}{l|}{Adamct} & 52.7 & \underline{1.309} & 5.73E-3 & 49.3 & \underline{1.177} & 1.3E-4 & 0 & \textbf{2} \\
        
        & \multicolumn{1}{l|}{LightGBM} & 55.0 & \textbf{1.331} & \underline{7.26E-3} & 51.0 & 0.890 & 1.4E-5 & \underline{1} & \underline{1} \\ 
        & \multicolumn{1}{l|}{DoubleEnsemble} & 54.0 & 1.225 & 5.75E-3 & - & - & - & 0 & 0\\
        & \multicolumn{1}{l|}{TCTS}           & 55.6 & 0.913 & 2.09E-3 & - & - & - & 0 & 0\\
        
    \midrule
        \multirow{3}*{\begin{tabular}{c}Time-series \\ Methods\end{tabular}}
        & \multicolumn{1}{l|}{iTransformer} & 53.8 & 1.095 & 4.01E-3 & - & - & - & 0 & 0\\
        & \multicolumn{1}{l|}{PatchTST}     & 53.0 & 1.274 & 5.17E-3 & - & - & - & 0 & 0\\
        & \multicolumn{1}{l|}{TimesNet}     & 55.5 & 1.131 & 6.16E-3 & - & - & - & 0 & 0\\
    \midrule
        \multirow{3}*{\begin{tabular}{c}Noisy Labels \\ Methods\end{tabular}}
        & \multicolumn{1}{l|}{CNLCU} & 52.6 & 1.233 & 5.03E-3 & 52.9 & \textbf{1.220}    & 1.2E-4 & \underline{1} & 0 \\ 
        & \multicolumn{1}{l|}{FINE}  & 55.3 & 1.070 & 4.35E-3 & \underline{56.3} & 0.863 & 9.7E-5 & 0 & \underline{1}\\ 
        & \multicolumn{1}{l|}{SEAL}  & \underline{56.6} & 1.200 & 5.95E-3 & 53.0 & 0.969 & 7.2E-5 & 0 & \underline{1} \\ 
    \midrule
        \multirow{3}*{Ours}
        & LA-Attention & \underline{56.6} & 1.142 & 5.27E-3 & 51.2 & 0.826 & 9.5E-5 & 0 & \underline{1}\\
        & RA-Labeling  & 55.2 & 1.038 & 3.53E-3 & 56.2 & 1.034 & \underline{1.4E-4} & 0 & \underline{1}\\
        & LARA         & \textbf{59.1} & 1.274 & \textbf{7.79E-3} & \textbf{57.8} & 1.059 & \textbf{1.5E-4} & \textbf{4} & 0\\
    \bottomrule
    \end{tabular}
    }
    \caption{Quantitative comparisons among different methods on the China's A-share stocks and the cryptocurrency (BTC/USDT). \emph{-} means that the corresponding method is either not implemented or unsuitable for corresponding setting. 
    We retrieve the top 1000 signals with the highest probability for each experiment. The best performance is highlighted in \textbf{bold}. The second best is highlighted with \underline{underline}.
    }
    \label{tab:exp-qlib-stock}
\end{table*}

In this section, we perform extensive experiments on three real-world financial markets: China’s A-share stock, the cryptocurrency~(OKEx), and ETFs to validate the effectiveness of LARA. 
We first introduce the experimental setup~(Sec.~\ref{sec:q1}), and then benchmark the LARA framework compared with deep learning methods on the Qlib quantitative investment platform~\cite{yang2020qlib}~(Sec.~\ref{sec:q2}). Finally, we provide experimental analysis to demonstrate the effectiveness of each component in LARA ~(Sec.~\ref{sec:q3}) and ablation studies to show the sensitivity of hyperparameters in LARA~(in Appendix).

\subsection{Experimental Setups}
\label{sec:q1}
\paragraph{Dataset Collection.}
\label{sec:exp-dataset}
For stocks, we use the dataset from Qlib: the daily data of the constituents of CSI300\footnote{CSI300 consists of the 300 largest and most liquid A-share stocks, which reflects the overall performance of the market.}. 
We follow the temporal order to split datasets into training~(01/01/2008--12/31/2014), validation~(01/01/2015--12/31/2016), and testing~(01/01/2017--08/01/2020) data. 
For cryptocurrency, we use the BTC/USDT trading pair. Each record is one market snapshot captured for approximately every 0.1 second. 
The training data comes from 7 consecutive trading days~(01/03/2021--07/03/2021), whose total quantity is 10 million. The validation and testing data come from the following 7 trading days~(08/03/2021--10/03/2021 and 11/03/2021--14/03/2021). 
We use technical factors~(\eg,~moving average of price) and limit-order book factors~(\eg,~buy/sell pressure indicators) to calculate input features~\cite{achelis2001technical}.
For ETFs, we conduct experiments over four Chinese exchange-traded funds~(ETFs): 159915.SZ, 512480.SH, 512880.SH, and 515050.SH. All of them are liquid and traded in large volumes. 
We chronologically split datasets into three time periods for training~(01/01/2020--17/04/2020), validation~(20/04/2020--29/05/2020), and testing~(01/06/2020--06/07/2020). 
Due to the space limit, we only show the experimental results of 512480.SH. 
The results for other ETFs, the complete factor list, and the statistics of all datasets can be found in Appendix.

\paragraph{Baselines.} 
(1)~\emph{Quantitative investment methods in Qlib:} 
ALSTM~\cite{qin2017dual}, 
TabNet~\cite{arik2020tabnet}, 
Transformer~\cite{vaswani2017attention}, 
Adamct~\cite{jiang2022adamct}, 
LightGBM~\cite{ke2017lightgbm},
DoubleEnsemble~\cite{zhang2020doubleensemble}, and 
TCTS~\cite{wu2021temp}. 
(2)~\emph{Time-series analysis methods:} Due to the temporal nature of daily stock data, we also implemented several SOTA time-series methods: 
iTransformer~\cite{liu2023itransformer}, 
PatchTST~\cite{nie2022time}, 
TimesNet\cite{wu2022timesnet}. 
(3)~\emph{Noisy labels methods:} 
CNLCU~\cite{xia2021sample}, 
FINE~\cite{kim2021fine}, 
SEAL~\cite{chen2021beyond}. 
For these methods, we utilize LightGBM as the base model.

\paragraph{Performance Metrics.}
\label{eva}
We use four commonly-used financial metrics~\cite{yang2020qlib} to evaluate the performance of LARA, \ie, \emph{Precision}~(PR), \emph{Win-Loss Ratio}~(WLR), \emph{\#Transactions}, and \emph{Average Return}~(AR). For the descriptions of each metric, please check the appendix for more details.

\paragraph{Evaluation Protocols.} 
Following \cite{zhang2020doubleensemble}, we adopt a representative trading strategy to evaluate the performance regarding the actual trading scenarios of the market. 
\emph{Note that we build individual models for long and short positions.}
(1) \emph{Trading signals:}
$\#$Transactions is an important criterion to measure the transaction frequency of trading strategies in high-frequency quantitative trading. According to \cite{abner2010etf}, it is a common practice to set thresholds to trigger tradings signals, but appropriate thresholds remains a challenging research issue. We retrieve the top 100 to 1000 samples with the highest predicted probability as the corresponding trading signals~(control the same trading frequency for different algorithms).
(2) \emph{Trading Strategies:}
we do not set a trading position limit. At a certain time step, if the model gives a positive signal, we open a new long position and close this position after the prediction horizon $\triangle$~(1 day/1 min/10 secs for stocks/ETFs/crypto). Similarly, we open short positions for the negative signals and close them after $\triangle$. 

For LARA, we use the default hyperparameter in \emph{LightGBM} and the \emph{metric learning module}. Other methods use their suggested parameters on Qlib or we conduct a gird search for them, the details can be found in Appendix.
All experiments are repeated 5 times, and we report the mean results with standard deviations. In what follows, we always record \emph{the average results of long and short positions.}

\begin{table}[tp]
    \centering
    \scalebox{0.84}{
    \begin{tabular}{cc|ccc}
    \toprule
        Training & Testing & PR~(\%) & WLR & AR~(\%)  \\
    \midrule
        - & - & $70.16_{\pm0.89}$ & $2.872_{\pm0.106}$  & $0.166_{\pm0.002}$ \\ 
        - & $\checkmark$ & $70.39_{\pm1.07}$ & $2.757_{\pm0.184}$ & $0.166_{\pm0.002}$ \\ 
        $\checkmark$ & - & $37.82_{\pm11.09}$ & $1.712_{\pm0.188}$ & $0.062_{\pm0.037}$ \\ 
        $\checkmark$ & $\checkmark$ & $\bm{78.58_{\pm0.67}}$ & $\bm{2.957_{\pm0.259}}$ & $\bm{0.196_{\pm0.003}}$ \\ 
    \bottomrule
    \end{tabular}
    }
    \caption{Ablation studies with LA-Attention applied in the training and testing phase, respectively. $\checkmark$(-) means that we evaluate w~(w/o) LA-Attention in the corresponding phase.}
    \label{tab:exp-subset}
\end{table}

\begin{figure*}[tbp]
	\centering
	\includegraphics[width=1.00\textwidth]{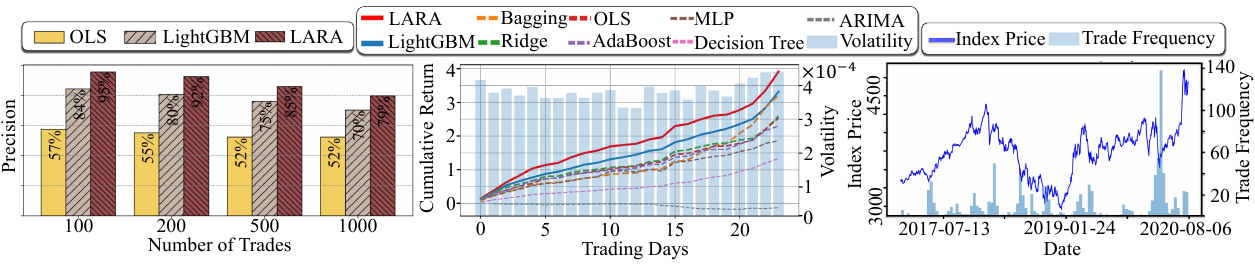}
	\caption{\textbf{Left:} Precision with the different number of trades~($\#$Transactions) among three methods on 512480.SH. \textbf{Middle:} Quantitative comparisons over cumulative return between LARA and a set of baselines on 512480.SH. The right y-axis illustrates the volatility \protect\cite{campbell1997econometrics} on each trading day. \textbf{Right:} Plot of the trade frequency in the China's A-share market and the corresponding stock index price. The left y-axis denotes the stock index price while the right y-axis denotes the trade frequency.}
	\label{fig:experiment-all}
\end{figure*}

\subsection{Experimental Results}
\label{sec:q2}
\paragraph{Main Results.}
In Table \ref{tab:exp-qlib-stock}, we show the quantitative results on China's A-share stocks and the cryptocurrency~(OKEx). 
LARA achieves 59.1\% and 57.8\% precision on stocks and crypto respectively, and outperforms the others by up to 2.5\% and 1.5\%.
As for the average return, LARA performs better than other competitors on stocks and improves a great margin on the crypto.
The outstanding performance of these two indicators demonstrates that LARA indeed has the ability to capture potential profitable signals.
For win-loss ratio, LARA can still achieve a decent effect comparable to other methods~(WLR is larger than 1), indicating a relatively low trading risk in real-world markets. In addition, noisy labels methods usually perform better than other methods, indicating the necessity of handling noisy labels on financial data. Our LARA not only extracts potential profitable samples but also dynamically modifies their labels for training, culminating in superior performance compared to other methods of noisy labels methods.

\paragraph{Effectiveness of LA-Attention.}
We then explore whether LA-Attention can boost performance when concentrating on the \emph{samples with high $p_\vx$}.
We conduct experiments on ETFs to consider whether to use LA-Attention in the training and testing phases.
In Table~\ref{tab:exp-subset}, we find that LA-Attention applied in both the training and testing phases outperforms other ablation competitors on all performance metrics.
This result is in line with the fact that LA-Attention can capture much more reliable trading opportunities with a significantly higher average return.
If we merely use LA-Attention in the [training/testing] phase, the performance decreases [8\%/40\%] on precision.
We posit that the significant gap stems from substantial deviations in the distributions of datasets between these two phases.
Thus, we should pay attention to the \emph{samples with high $p_\vx$} in both two phases to capture more reliable trading opportunities.

\subsection{Experimental Analysis}
\label{sec:q3}
\paragraph{Transactions. }
First, in order to investigate the effect of $\#$Transactions, we choose two representative methods: OLS, LightGBM~(widely used for investment).
In Fig.~\ref{fig:experiment-all} left, LARA consistently outperforms the other two models in terms of precision under different $\#$Transactions. 
These results highlight the effectiveness of LARA across different trading frequencies.

\paragraph{Trading Performance. }
Second, we illustrate the performance of the back-testing strategy in terms of cumulative return~(Fig.~\ref{fig:experiment-all} middle). We choose the top 1000 signals with the highest predicted probability for fair comparison and do not set position limits. 
For the sake of simplicity, transaction costs are ignored since $\#$Transaction is the same over all methods, which does not change conclusions.
It is obvious that LARA consistently outperforms other baseline models throughout the testing period, which demonstrates that LARA indeed improves the prediction ability of the model on the asset price trend.

\paragraph{Trade Distribution. }
Third, in Fig.~\ref{fig:experiment-all} right, we plot the trade distribution of LARA in China’s A-share market associated with the market index price.
We can clearly see that when the index price fluctuates sharply, LARA trades stocks in a higher frequency. It meets our expectation that there are more opportunities when the market is highly volatile. These results further demonstrate that LARA can be used to guide real-world quantitative investments to pick up reliable trading opportunities.

\section{Conclusion}
\label{sec:conclusion}
We study the problem of price movement forecasting and propose the LARA (\underline{L}ocality-Aware \underline{A}ttention and Iterative \underline{R}efinement L\underline{a}beling) framework. In LARA, we introduce LA-Attention to extract potentially profitable samples and RA-Labeling to adaptively refine the noisy labels of potentially profitable samples. 
Extensive experiments on three real-world financial markets showcase its superior performance over time-series analysis methods, machine learning based models and noisy labels methods.
Besides, we also illustrate how each component works by comprehensive ablation studies, which indicates that LARA indeed captures more reliable trading opportunities.
We expect our work to pave the way for price movement forecasting in more realistic quantitative trading scenarios.

\section*{Acknowledgements}
Jian Li, Liang Zeng, Lei Wang and Hui Niu were supported in part by the National Natural Science Foundation of China Grant 62161146004.

\section*{Contribution Statement}
Liang Zeng and Lei Wang have made equal and significant contributions to this work, including the methods, implementation, and paper writing. 
Hui Niu conducted the literature review and implemented the evaluation framework.
Ruchen Zhang and Ling Wang provided the source data to support this work.
Jian Li, as the corresponding author, contributed to the idea and paper writing, as well as providing computing resources.

\bibliographystyle{named}
\bibliography{ijcai24}

\begin{thebibliography}{}

\bibitem[\protect\citeauthoryear{Abner}{2010}]{abner2010etf}
David~J Abner.
\newblock {\em The ETF handbook: how to value and trade exchange traded funds}.
\newblock John Wiley \& Sons, 2010.

\bibitem[\protect\citeauthoryear{Achelis}{2001}]{achelis2001technical}
Steven~B Achelis.
\newblock {\em Technical Analysis from A to Z}.
\newblock McGraw Hill New York, 2001.

\bibitem[\protect\citeauthoryear{Ar{\i}k and Pfister}{2021}]{arik2020tabnet}
Sercan~O Ar{\i}k and Tomas Pfister.
\newblock Tabnet: Attentive interpretable tabular learning.
\newblock In {\em Proc. of AAAI}, 2021.

\bibitem[\protect\citeauthoryear{Bloomfield \bgroup \em et al.\egroup }{2009}]{bloomfield2009noise}
Robert Bloomfield, Maureen O’hara, and Gideon Saar.
\newblock How noise trading affects markets: An experimental analysis.
\newblock {\em The Review of Financial Studies}, 2009.

\bibitem[\protect\citeauthoryear{Campbell \bgroup \em et al.\egroup }{1997}]{campbell1997econometrics}
John~Y Campbell, Andrew~W Lo, A~Craig MacKinlay, and Robert~F Whitelaw.
\newblock {\em The econometrics of financial markets}.
\newblock Princeton University Press, 1997.

\bibitem[\protect\citeauthoryear{Chen \bgroup \em et al.\egroup }{2021}]{chen2021beyond}
Pengfei Chen, Junjie Ye, Guangyong Chen, Jingwei Zhao, and Pheng-Ann Heng.
\newblock Beyond class-conditional assumption: A primary attempt to combat instance-dependent label noise.
\newblock In {\em Proceedings of the AAAI Conference on Artificial Intelligence}, volume~35, pages 11442--11450, 2021.

\bibitem[\protect\citeauthoryear{De~Prado}{2018}]{de2018advances}
Marcos~Lopez De~Prado.
\newblock {\em Advances in financial machine learning}.
\newblock John Wiley \& Sons, 2018.

\bibitem[\protect\citeauthoryear{De~Vazelhes \bgroup \em et al.\egroup }{2020}]{de2020metric}
William De~Vazelhes, CJ~Carey, Yuan Tang, Nathalie Vauquier, and Aur{\'e}lien Bellet.
\newblock metric-learn: Metric learning algorithms in python.
\newblock {\em Journal of Machine Learning Research}, 2020.

\bibitem[\protect\citeauthoryear{Fama}{1960}]{fama1960efficient}
Eugene~F Fama.
\newblock Efficient market hypothesis.
\newblock {\em Diss. PhD Thesis, Ph. D. dissertation}, 1960.

\bibitem[\protect\citeauthoryear{Grossman and Stiglitz}{1980}]{grossman1980impossibility}
Sanford~J Grossman and Joseph~E Stiglitz.
\newblock On the impossibility of informationally efficient markets.
\newblock {\em The American economic review}, 1980.

\bibitem[\protect\citeauthoryear{Gu \bgroup \em et al.\egroup }{2018}]{gu2018empirical}
Shihao Gu, Bryan Kelly, and Dacheng Xiu.
\newblock Empirical asset pricing via machine learning.
\newblock Technical report, 2018.

\bibitem[\protect\citeauthoryear{Harvey \bgroup \em et al.\egroup }{2016}]{harvey2016and}
Campbell~R Harvey, Yan Liu, and Heqing Zhu.
\newblock … and the cross-section of expected returns.
\newblock {\em The Review of Financial Studies}, 2016.

\bibitem[\protect\citeauthoryear{Jiang \bgroup \em et al.\egroup }{2022}]{jiang2022adamct}
Juyong Jiang, Jae~Boum Kim, Yingtao Luo, Kai Zhang, and Sunghun Kim.
\newblock Adamct: Adaptive mixture of cnn-transformer for sequential recommendation.
\newblock {\em arXiv:2205.08776}, 2022.

\bibitem[\protect\citeauthoryear{Kaya and Bilge}{2019}]{kaya2019deep}
Mahmut Kaya and Hasan~{\c{S}}akir Bilge.
\newblock Deep metric learning: A survey.
\newblock {\em Symmetry}, 11(9):1066, 2019.

\bibitem[\protect\citeauthoryear{Ke \bgroup \em et al.\egroup }{2017}]{ke2017lightgbm}
Guolin Ke, Qi~Meng, Thomas Finley, Taifeng Wang, Wei Chen, Weidong Ma, Qiwei Ye, and Tie-Yan Liu.
\newblock Lightgbm: A highly efficient gradient boosting decision tree.
\newblock {\em NeurIPS}, 2017.

\bibitem[\protect\citeauthoryear{Kim \bgroup \em et al.\egroup }{2021}]{kim2021fine}
Taehyeon Kim, Jongwoo Ko, JinHwan Choi, Se-Young Yun, et~al.
\newblock Fine samples for learning with noisy labels.
\newblock {\em Advances in Neural Information Processing Systems}, 34:24137--24149, 2021.

\bibitem[\protect\citeauthoryear{Li \bgroup \em et al.\egroup }{2018}]{livisualizing}
Hao Li, Zheng Xu, Gavin Taylor, Christoph Studer, and Tom Goldstein.
\newblock Visualizing the loss landscape of neural nets.
\newblock In {\em Proc. of NeurIPS}, 2018.

\bibitem[\protect\citeauthoryear{Li \bgroup \em et al.\egroup }{2019}]{li2019individualized}
Zhige Li, Derek Yang, Li~Zhao, Jiang Bian, Tao Qin, and Tie-Yan Liu.
\newblock Individualized indicator for all: Stock-wise technical indicator optimization with stock embedding.
\newblock In {\em Proc. of KDD}, 2019.

\bibitem[\protect\citeauthoryear{Liu \bgroup \em et al.\egroup }{2023}]{liu2023itransformer}
Yong Liu, Tengge Hu, Haoran Zhang, Haixu Wu, Shiyu Wang, Lintao Ma, and Mingsheng Long.
\newblock itransformer: Inverted transformers are effective for time series forecasting.
\newblock {\em arXiv preprint arXiv:2310.06625}, 2023.

\bibitem[\protect\citeauthoryear{Lo and MacKinlay}{2011}]{lo2011non}
Andrew~W Lo and A~Craig MacKinlay.
\newblock {\em A non-random walk down Wall Street}.
\newblock Princeton University Press, 2011.

\bibitem[\protect\citeauthoryear{Lo}{2019}]{lo2019adaptive}
Andrew~W Lo.
\newblock {\em Adaptive markets: Financial evolution at the speed of thought}.
\newblock Princeton University Press, 2019.

\bibitem[\protect\citeauthoryear{Nie \bgroup \em et al.\egroup }{2022}]{nie2022time}
Yuqi Nie, Nam~H Nguyen, Phanwadee Sinthong, and Jayant Kalagnanam.
\newblock A time series is worth 64 words: Long-term forecasting with transformers.
\newblock In {\em The Eleventh International Conference on Learning Representations}, 2022.

\bibitem[\protect\citeauthoryear{Qin \bgroup \em et al.\egroup }{2017}]{qin2017dual}
Yao Qin, Dongjin Song, Haifeng Chen, Wei Cheng, Guofei Jiang, and Garrison~W Cottrell.
\newblock A dual-stage attention-based recurrent neural network for time series prediction.
\newblock In {\em Proc. of IJCAI}, 2017.

\bibitem[\protect\citeauthoryear{Sawhney \bgroup \em et al.\egroup }{2021}]{sawhney2021exploring}
Ramit Sawhney, Shivam Agarwal, Arnav Wadhwa, and Rajiv Shah.
\newblock Exploring the scale-free nature of stock markets: Hyperbolic graph learning for algorithmic trading.
\newblock In {\em Proc. of WWW}, 2021.

\bibitem[\protect\citeauthoryear{Song \bgroup \em et al.\egroup }{2022}]{song2022learning}
Hwanjun Song, Minseok Kim, Dongmin Park, Yooju Shin, and Jae-Gil Lee.
\newblock Learning from noisy labels with deep neural networks: A survey.
\newblock {\em IEEE TNNLS}, 2022.

\bibitem[\protect\citeauthoryear{Van~der Maaten and Hinton}{2008}]{van2008visualizing}
Laurens Van~der Maaten and Geoffrey Hinton.
\newblock Visualizing data using t-sne.
\newblock {\em JMLR}, 2008.

\bibitem[\protect\citeauthoryear{Vaswani \bgroup \em et al.\egroup }{2017}]{vaswani2017attention}
Ashish Vaswani, Noam Shazeer, Niki Parmar, Jakob Uszkoreit, Llion Jones, Aidan~N Gomez, {\L}ukasz Kaiser, and Illia Polosukhin.
\newblock Attention is all you need.
\newblock In {\em Proc. of NeurIPS}, 2017.

\bibitem[\protect\citeauthoryear{Wang \bgroup \em et al.\egroup }{2019a}]{wang2019clvsa}
Jia Wang, Tong Sun, Benyuan Liu, Yu~Cao, and Hongwei Zhu.
\newblock Clvsa: A convolutional lstm based variational sequence-to-sequence model with attention for predicting trends of financial markets.
\newblock In {\em Proc. of IJCAI}, 2019.

\bibitem[\protect\citeauthoryear{Wang \bgroup \em et al.\egroup }{2019b}]{wang2019alphastock}
Jingyuan Wang, Yang Zhang, Ke~Tang, Junjie Wu, and Zhang Xiong.
\newblock Alphastock: A buying-winners-and-selling-losers investment strategy using interpretable deep reinforcement attention networks.
\newblock In {\em Proc. of KDD}, 2019.

\bibitem[\protect\citeauthoryear{Wu \bgroup \em et al.\egroup }{2021}]{wu2021temp}
Xueqing Wu, Lewen Wang, Yingce Xia, Weiqing Liu, Lijun Wu, Shufang Xie, Tao Qin, and Tie-Yan Liu.
\newblock Temporally correlated task scheduling for sequence learning.
\newblock In {\em Proc. of ICML}, 2021.

\bibitem[\protect\citeauthoryear{Wu \bgroup \em et al.\egroup }{2022}]{wu2022timesnet}
Haixu Wu, Tengge Hu, Yong Liu, Hang Zhou, Jianmin Wang, and Mingsheng Long.
\newblock Timesnet: Temporal 2d-variation modeling for general time series analysis.
\newblock In {\em The Eleventh International Conference on Learning Representations}, 2022.

\bibitem[\protect\citeauthoryear{Xia \bgroup \em et al.\egroup }{2021}]{xia2021sample}
Xiaobo Xia, Tongliang Liu, Bo~Han, Mingming Gong, Jun Yu, Gang Niu, and Masashi Sugiyama.
\newblock Sample selection with uncertainty of losses for learning with noisy labels.
\newblock In {\em International Conference on Learning Representations}, 2021.

\bibitem[\protect\citeauthoryear{Xing \bgroup \em et al.\egroup }{2002}]{xing2002distance}
Eric Xing, Michael Jordan, Stuart~J Russell, and Andrew Ng.
\newblock Distance metric learning with application to clustering with side-information.
\newblock In {\em Proc. of NeurIPS}, 2002.

\bibitem[\protect\citeauthoryear{Xu \bgroup \em et al.\egroup }{2020}]{xu2020adaptive}
Jin Xu, Jingbo Zhou, Yongpo Jia, Jian Li, and Xiong Hui.
\newblock An adaptive master-slave regularized model for unexpected revenue prediction enhanced with alternative data.
\newblock In {\em Proc. of ICDE}, 2020.

\bibitem[\protect\citeauthoryear{Yang \bgroup \em et al.\egroup }{2020}]{yang2020qlib}
Xiao Yang, Weiqing Liu, Dong Zhou, Jiang Bian, and Tie-Yan Liu.
\newblock Qlib: An ai-oriented quantitative investment platform.
\newblock {\em arXiv:2009.11189}, 2020.

\bibitem[\protect\citeauthoryear{Zhang \bgroup \em et al.\egroup }{2017}]{zhang2017stock}
Liheng Zhang, Charu Aggarwal, and Guo-Jun Qi.
\newblock Stock price prediction via discovering multi-frequency trading patterns.
\newblock In {\em Proc. of KDD}, 2017.

\bibitem[\protect\citeauthoryear{Zhang \bgroup \em et al.\egroup }{2020}]{zhang2020doubleensemble}
Chuheng Zhang, Yuanqi Li, Xi~Chen, Yifei Jin, Pingzhong Tang, and Jian Li.
\newblock Doubleensemble: A new ensemble method based on sample reweighting and feature selection for financial data analysis.
\newblock In {\em Proc. of ICDM}, 2020.

\end{thebibliography}


\end{document}